\documentclass[11pt,a4paper]{article}
\usepackage[hyperref]{emnlp2018}
\usepackage{times}
\usepackage{latexsym}
\usepackage{tikz}
\usepackage{amsmath}
\usepackage{arydshln}
\usepackage{multirow}
\usepackage[linesnumbered,ruled,vlined]{algorithm2e}
\usetikzlibrary{positioning}

\usepackage{url}

\aclfinalcopy 

\title{An AMR Aligner Tuned by Transition-based Parser}

\author{Yijia Liu, Wanxiang Che\thanks{* Email corresponding.}, Bo Zheng, Bing Qin, Ting Liu \\
	Research Center for Social Computing and Information Retrieval \\
	Harbin Institute of Technology, China \\
	{\tt \{yjliu,car,bzheng,qinb,tliu\}@ir.hit.edu.cn}	
}

\date{}

\begin{document}
\maketitle
\begin{abstract}
In this paper, we propose a new rich resource
enhanced AMR aligner which produces multiple
alignments and a new transition system for AMR parsing
along with its oracle parser.
Our aligner is further tuned by
our oracle parser via picking the alignment that
leads to the highest-scored achievable AMR graph.
Experimental results show that our aligner outperforms the rule-based aligner in previous work
by achieving higher alignment F1 score and consistently improving two open-sourced AMR parsers.
Based on our aligner and transition system, 
we develop a transition-based AMR parser that parses a sentence into its AMR graph directly.
An ensemble of our parsers with only words and POS tags as input
leads to 68.4 Smatch F1 score,
which outperforms the parser of \citet{wang-xue:2017:EMNLP2017}.
\end{abstract}

\section{Introduction}\label{sec:intro}

\begin{figure}[t]
	\includegraphics[width=\columnwidth, trim={0, 1cm, 0, 1cm}, clip]{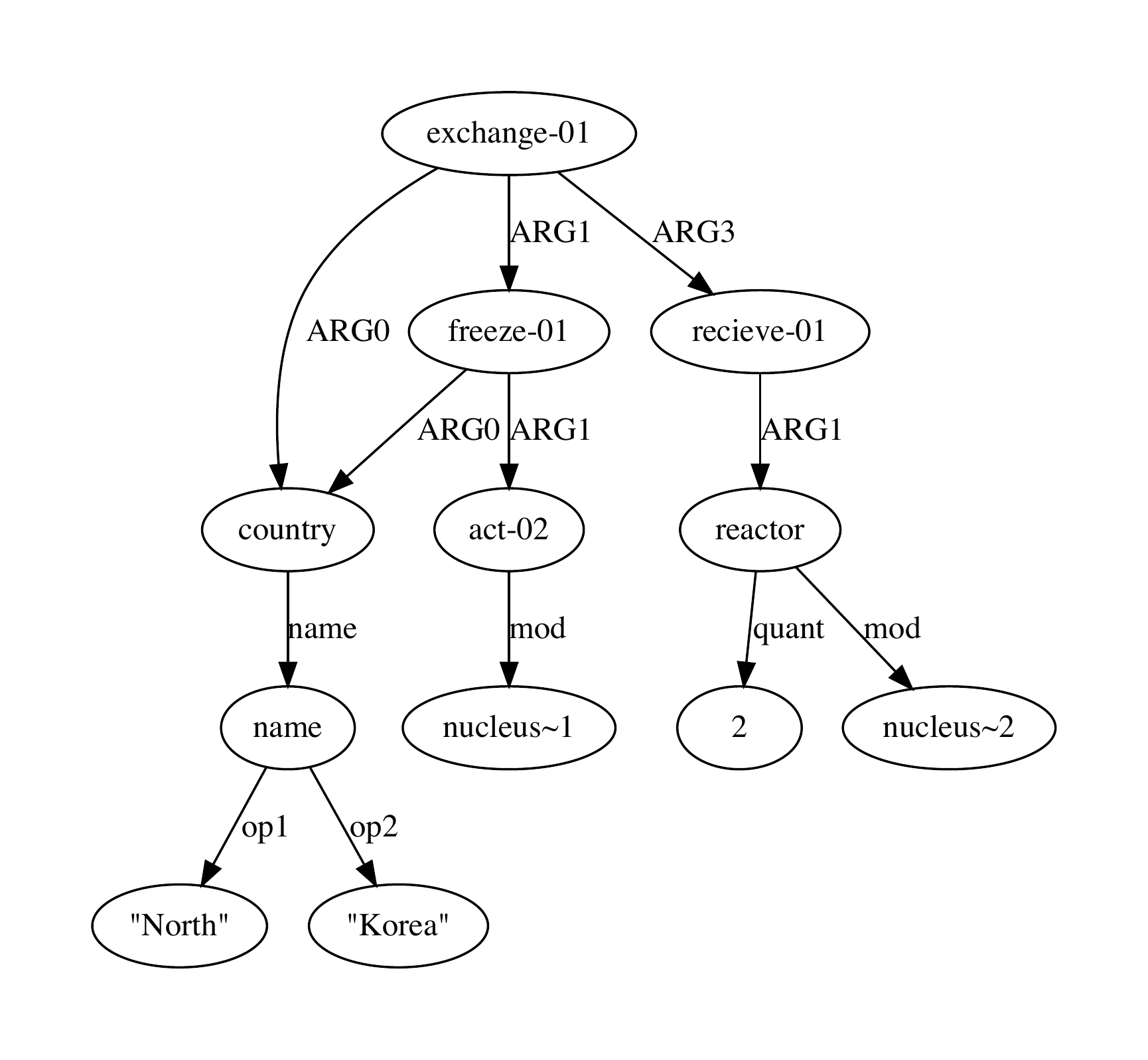}
	\caption{AMR graph for the sentence \textit{``North Korea froze its nuclear actions in exchange for two nuclear reactors.''}}
	\label{fig:eg_amr}
\end{figure}

Abstract Meaning Representation (AMR) \cite{banarescu-EtAl:2013:LAW7-ID}
is a semantic representation which encodes the meaning of a sentence
in a rooted and directed graph, 
whose nodes are abstract semantic concepts and edges are semantic relations between concepts
(see Figure \ref{fig:eg_amr} for an example).
Parsing a sentence into its AMR graph has
drawn a lot of research attention in recent years
with a number of parsers being developed
\cite{flanigan-EtAl:2014:P14-1,
	wang-xue-pradhan:2015:NAACL-HLT,
	pust-EtAl:2015:EMNLP,
	artzi-lee-zettlemoyer:2015:EMNLP,
	peng-song-gildea:2015:CoNLL,
	zhou-EtAl:2016:EMNLP20163,
	goodman-vlachos-naradowsky:2016:P16-1,
	damonte-cohen-satta:2017:EACLlong,
	ballesteros-alonaizan:2017:EMNLP2017,
	foland-martin:2017:Long,
	konstas-EtAl:2017:Long}.

The nature of abstracting away the association between a concept 
and a span of words complicates the training of the AMR parser.
A word-concept aligner is required to
derive such association from the sentence-AMR-graph pair and 
the alignment output is then used as reference to train the AMR parser.
In previous works, such alignment is extracted
by either greedily applying a set of heuristic rules
\citep{flanigan-EtAl:2014:P14-1}
or adopting the unsupervised word alignment technique from
machine translation \citep{pourdamghani-EtAl:2014:EMNLP2014,wang-xue:2017:EMNLP2017}.

The rule-based aligner --- JAMR aligner proposed by \citet{flanigan-EtAl:2014:P14-1}
is widely used in previous works 
thanks to its flexibility of incorporating additional linguistic resources
like WordNet.
However, achieving good alignments with the JAMR aligner still faces some difficult challenges.
The first challenge is deriving an optimal alignment in ambiguous situations.
Taking the sentence-AMR-graph pair in Figure \ref{fig:eg_amr} for example,
the JAMR aligner doesn't distinguish between the two ``{\it nuclear}''s in the sentence
and can yield sub-optimal alignment in which the first ``{\it nuclear}'' is aligned
to the {\tt nucleus\url{~}2} concept.\footnote{We use the tail number in {\tt nucleus\url{~}X}
to distinguish two different concepts in the AMR graph.
We need to note that there is no {\tt nucleus\url{~}2} in the AMR corpus.}
The second challenge is recalling more semantically matched word-concept pair
without harming the alignment precision.
The JAMR aligner adopts a rule that aligns the word-concept pair
which at least have a common longest prefix of 4 characters,
but omitting the shorter cases like aligning the word ``\textit{actions}'' to the concept \texttt{act-01}
and the semantically matched cases like aligning the word ``\textit{example}'' to the concept \texttt{exemplify-01}.
The final challenge which is faced by both the rule-based and unsupervised aligners
is tuning the alignment with downstream parser learning.
Previous works treated the alignment as a fixed input.
Its quality is never evaluated and its alternatives are never explored.
All these challenges make the JAMR aligner achieve only an alignment F1 score
of about 90\% and
influence the performance of the trained AMR parsers.

In this paper, we propose a novel method to solve these challenges and
improve the word-to-concept alignment,
which further improves the AMR parsing performance.
A rule-based aligner and 
a transition-based oracle AMR parser
lie in the core of our method.
For the aligner part, we
incorporate rich semantic resources into the JAMR aligner
to recall more word-concept pairs
and cancel its greedily aligning process. 
This leads to multiple alignment outputs with higher recall but lower precision.
For the parser part,
we propose a new transition system that can parse the raw
sentence into AMR graph directly.
Meanwhile, a new oracle algorithm is proposed which
produces the best achievable AMR graph from an alignment.
Our aligner is tuned by our oracle parser by feeding the alignments
to the oracle parser and picking the one which leads to
the highest Smatch F1 score \cite{cai-knight:2013:Short}.
The chosen alignment is used in downstream training of the AMR parser.
Based on the newly proposed aligner and transition system,
we develop a transition-based parser that directly parses a sentence
into its AMR graph and it can be easily improved through ensemble
thanks to its simplicity.

We conduct experiments on LDC2014T12 dataset.\footnote{\url{catalog.ldc.upenn.edu/ldc2014t12}}
Both intrinsic and extrinsic evaluations are performed on our aligner.
In the intrinsic evaluation, our aligner achieves an alignment F1 score of 95.2\%.
In the extrinsic evaluation, we replace the JAMR aligner with ours
in two open-sourced AMR parsers, which leads to consistent improvements on both parsers.
We also evaluate our transition-based parser on the same dataset.
Using both our aligner and ensemble, a score of
68.1 Smatch F1 is achieved
without any additional resources, which is comparable to the parser of \citet{wang-xue:2017:EMNLP2017}.
With additional part-of-speech (POS) tags,
our ensemble parser
achieves 68.4 Smatch F1 score and
outperforms that of \citet{wang-xue:2017:EMNLP2017}.

The contributions of this paper come in two folds:

\begin{itemize}
	\item We propose a new AMR aligner (\S\ref{sec:aligner}) which recalls 
	more semantically matched pairs and produces multiple alignments.
	We also propose a new transition system for AMR parsing (\S\ref{sec:trans-sys}) and 
	use its oracle (\S\ref{sec:oracle-parser})
	to pick the alignment that leads to the highest-scored achievable AMR graph (\S\ref{sec:tune}).
	Both intrinsic and extrinsic evaluations (\S\ref{sec:align-exp}) show the effectiveness of our aligner
	by achieving higher F1 score and
	consistently improving two open-sourced AMR parsers.
	
	\item We build a new transition-based parser (\S\ref{sec:parser}) upon our aligner and transition system
	which directly parses a raw sentence into its AMR graph.
	Through simple ensemble, our parser achieves 68.4 Smatch F1 score with
	only words and POS tags as input (\S\ref{sec:parse-exp}) and outperforms the parser of \citet{wang-xue:2017:EMNLP2017}.
	
\end{itemize}

Our code and the alignments for LDC2014T12 dataset are publicly available at \url{https://github.com/Oneplus/tamr}

\section{Related Work}

\paragraph{AMR Parsers.}
AMR parsing maps a natural language sentence into
its AMR graph.
Most current parsers construct the AMR graph in a two-staged manner
which first identifies concepts (nodes in the graph) from the
input sentence, then identifies relations (edges in the graph)
between the identified concepts.
\citet{flanigan-EtAl:2014:P14-1} and
their follow-up works \cite{flanigan-EtAl:2016:SemEval,zhou-EtAl:2016:EMNLP20163} model
the parsing problem as finding the maximum spanning connected graph.
\citet{wang-xue-pradhan:2015:NAACL-HLT} proposes to greedily transduce
the dependency tree into AMR graph and a bunch of works
\citep{wang-xue-pradhan:2015:ACL-IJCNLP,goodman-vlachos-naradowsky:2016:P16-1,wang-xue:2017:EMNLP2017}
further improve the transducer's performance with rich features and
imitation learning.\footnote{\citet{wang-xue-pradhan:2015:NAACL-HLT} and the follow-up works refer their transducing process
	as ``transition-based''. However, to distinguish their work with that of \citet{damonte-cohen-satta:2017:EACLlong} and \citet{ballesteros-alonaizan:2017:EMNLP2017},
	we use the term ``transduce'' instead.}
Transition-based methods that directly parse an input sentence
into its AMR graph have also been studied
\cite{ballesteros-alonaizan:2017:EMNLP2017,damonte-cohen-satta:2017:EACLlong}.
In these works, the concept identification and relation identification 
are performed jointly. 

An aligner which maps a span of words into its concept 
serves to the generation of training data for the concept identifier,
thus is important to the parser training.
Missing or incorrect alignments lead to poor concept identification,
which then hurt the overall AMR parsing performance.
Besides the typical two-staged methods,
the aligner also works
in some other AMR parsing algorithms like 
that using syntax-based machine translation \cite{pust-EtAl:2015:EMNLP},
sequence-to-sequence \cite{peng-EtAl:2017:EACLlong1,konstas-EtAl:2017:Long},
Hyperedge Replacement Grammar
\cite{peng-song-gildea:2015:CoNLL} and
Combinatory Category Grammar \cite{artzi-lee-zettlemoyer:2015:EMNLP}.
\citet{P18-1171} proposed a transition-based parser
that parses AMR graph from the identified concepts,
which is close to the transition-based method in terms of .
However, their work relies on the identified concepts
and can be treated as a two-staged method.

Previous aligner works solve the alignment problem
in two different ways.
The rule-based aligner \cite{flanigan-EtAl:2014:P14-1}
defines a set of heuristic rules which align a span of
words to the graph fragment and greedily applies
these rules.
The unsupervised aligner
\cite{pourdamghani-EtAl:2014:EMNLP2014,wang-xue:2017:EMNLP2017}
uncovers the word-to-concept alignment from
the linearized AMR graph through EM.
All these approaches yield a single alignment for one sentence
and its effect on the downstream parsing is not considered.
\citet{P18-1037} proposed to
model alignment as a latent variable.
Improved performance was achieved
which indicates the usefulness of improving alignment.
In addition to the work on AMR-to-string alignment,
there are also works trying to uncover the 
alignment between AMR  and dependency syntax \citep{szubert-lopez-schneider:2018:N18-1}.

\paragraph{JAMR Aligner \cite{flanigan-EtAl:2014:P14-1}.}
Two components exist in the JAMR aligner:
1) a set of heuristic rules and 2) a greedy search process.

The heuristic rules in the JAMR aligner
are a set of indicator functions $\rho(c, w_{s, e})$
which take a concept $c$ and a span of words $w_{s, e}$ starting from $s$ and ending with $e$
as input and return whether
they should be aligned.
These rules  can be categorized into {\it matching rules} and {\it updating rules}.
The matching rules directly compare $c$ with $w_{s, e}$ and determine if they should be aligned.
The updating rules first retrieve the concept $c'$ that $w_{s, e}$ aligns,
then determine if $c$ and $w_{s, e}$ should be aligned by checking whether
$c$ and $c'$ meet some conditions.
Here, we illustrate how \textit{update rules} work by applying a rule named \textbf{Entity Type}
on the AMR graph in Figure \ref{fig:eg_amr}
as an example.
When determining if the entity type concept \texttt{country} should be aligned to ``\textit{North Korea}'',
the \textbf{Entity Type} rule first retrieve that this span is aligned to the fragment \texttt{(name :op1 "North" :op2 "Korea")},
then determine if they are aligned by checking if \texttt{name} is the tail concept of \texttt{country}.

The greedy search process applies rules in a manually defined order.
The results are mutually exclusive which means
once a graph fragment is aligned by one rule, it cannot be realigned.
By doing so, conflicts between the alignments produced by 
different rules are resolved.
\citet{flanigan-EtAl:2014:P14-1} didn't talk about the principle of orders
but it generally follows the principle that 1) the matching rules have higher priorities
than the updating rules, 
and 2) exact matching rules have higher priorities than the fuzzy matching rules.

\section{Enhanced Rule-based Aligner}\label{sec:aligner}

\subsection{Enhancing Aligner with Rich Semantic Resources}

Error propagates in the greedy search process.
An alignment error can lead to future errors because of the
dependencies and mutual exclusions between rules.
In the JAMR aligner, rules that recall more
alignments but introduce errors are carefully
opted out and it influences the aligner's performance.
Our motivation is to use rich semantic resources
to recall more alignments.
Instead of resolving the resulted conflicts and errors by greedy search,
we keep the multiple alignments produced by the aligner
and let a parser decide the best alignment.

In this paper, we use two kinds of semantic resources to recall more
alignments, which include the similarity drawn
from \textit{Glove embedding} \cite{pennington-socher-manning:2014:EMNLP2014}\footnote{\url{nlp.stanford.edu/projects/glove/}} and
the {\it morphosemantic} database \cite{10.1007/978-3-642-04235-5_30} in the WordNet project\footnote{\url{wordnet.princeton.edu/wordnet/download/standoff/}}.
Two additional matching schemes {\bf semantic match} and {\bf morphological match} are proposed as:

\paragraph{Semantic Match.} \textit{Glove embedding}
encodes a word into its vector representation.
We define \textit{semantic match} of a concept
as a word in the sentence
that has a cosine similarity greater than 0.7 in the embedding space
with the concept striping off
trailing number (e.g. {\tt run-01} $\rightarrow$ {\tt run}).

\paragraph{Morphological Match.} \textit{Morphosemantic} is a 
database that contains links among derivational links connecting
noun and verb senses (e.g., ``{\it example}'' and {\tt exemplify}).
We define morphological match of a concept as a word in the sentence 
having the ({\it word}, \texttt{concept}) link in the database.

\begin{table}[t]
	\centering
	\small
	\begin{tabular}{p{0.95\columnwidth}}
		{\bf (Semantic Named Entity)} Applies to {\tt name} concepts and
		their {\tt opn} children. Matches a span that matches 
		the semantic match of each child in numerical order.\\[0.3em]
		{\bf (Morphological Named Entity)} Applies to {\tt name} concepts and
		their {\tt opn} children. Matches a span that matches 
		the morphological match of each child in numerical order.\\[0.3em]
		{\bf (Semantic Concept)} Applies to any concept. 
		Strips off trailing `-[0-9]+' from the concept,
		and matches any semantic matching word. \\[0.3em]
		{\bf (Morphological Concept)} Applies to any concept. 
		Strips off trailing `-[0-9]+' from the concept,
		and matches any morphological matching word or WordNet lemma. \\
	\end{tabular}
	\caption{The extended rules.} \label{tbl:extend-rules}
\end{table}
By defining the {\bf semantic match} and {\bf morphological match},
we extend the rules in \citet{flanigan-EtAl:2014:P14-1} with four
additional matching rules as shown in Table \ref{tbl:extend-rules}.
These rules are intended to recall the concepts or entities which
either semantically resemble a span of words but differ in the surface form,
or match a span of words in their morphological derivation.

\subsection{Producing Multiple Alignments}
\begin{algorithm}[t]
	\small
	\SetKwComment{Comment}{$\triangleright$\ }{}
	\KwIn{An AMR graph with a set of graph fragments $C$;
		a sentence $W$; a set of matching rules $\mathrm{P}_{M}$;
		and a set of updating rules $\mathrm{P}_U$.}
	\KwOut{a set of alignments $\mathcal{A}$.}
	\For{$c \in C$} {\label{algo:align:init_begin}
		$A_c \gets \emptyset$\;
	}\label{algo:align:init_end}
	\For{$\rho_M \in \mathrm{P}_{M}$}{\label{algo:align:match_begin}
		\For{$w_{s, e} \gets$ \textsf{spans}($W$)}{
			\For{$c \in C$} {
				\If{$\rho_M(c, w_{s, e})$}{
					$A_c \gets A_c \cup (s, e, \textsf{nil})$\;
				}
			}
		}
	}\label{algo:align:match_end}
	{\it updated} $\gets$ \textsf{true} \;\label{algo:align:update_begin}
	\While{\text{updated} is \textsf{true}} {
		{\it updated} $\gets$ \textsf{false}\;
		\For {$\rho_U \in \mathrm{P}_U$} {
			\For{$c, c' \in C \times C$} {
				\For{$(s, e, d) \in A_c'$}{
					\If{$\rho_U(c, w_{s, e}) \land (s, e, c') \notin A_c$}{
						$A_c \gets A_c \cup (s, e, c')$\;
						{\it updated} $\gets$ \textsf{true};
					}
				}
			}
		}
	}\label{algo:align:update_end}
	$\mathcal{A} \gets \emptyset$ \;\label{algo:align:enumerate_begin}
	\For{$(a_1, ..., a_c) \in \textsf{CartesianProduct}(A_1, ..., A_{|C|})$} {
		{\it legal} $\gets$ \textsf{true}\;
		\For{$a \in (a_1, ..., a_c)$}{
			$(s, e, c') \gets a$\;
			$(s', e', d) \gets a_{c'}$\;
			\If{$s \neq s' \land e \neq e'$}{
				{\it legal} $\gets$ \textsf{false} \;
			}
		}
		\If{legal}{
			$\mathcal{A} \gets \mathcal{A} \cup (a_1, ..., a_c)$\;
		}
	}\label{algo:align:enumerate_end}
	\caption{Our alignment algorithm.}\label{algo:align}
\end{algorithm}

Using the rules in the JAMR aligner along with our four extended matching rules,
we propose an algorithm to draw multiple alignments 
from a pair of sentence and AMR graph and it is shown in Algorithm \ref{algo:align}.
In this algorithm, $A_c$ denotes the set of candidate alignments for a graph fragment $c$,
in which each alignment is represented as a tuple $(s, e, c')$
where $s$ denotes the starting position, $e$ denotes the ending position,
and $c'$ denotes the concept that lead to this alignment.
At the beginning, $A_c$ is initialized as an empty set (line \ref{algo:align:init_begin} to \ref{algo:align:init_end}).
Then all the matching rules are tried to align a span of words to that fragment
(line \ref{algo:align:match_begin} to \ref{algo:align:match_end}).
After applying all the matching rules, all the updating rules are repeatedly
applied until no new alignment is generated in one iteration (line \ref{algo:align:update_begin} to \ref{algo:align:update_end}).
During applying the updating rules, we keep track of the dependencies between fragments.
Finally, all the possible combination of the alignments
are enumerated without considering the one that violates the fragment dependencies
(line \ref{algo:align:enumerate_begin} to \ref{algo:align:enumerate_end}). 

\section{Transition-based AMR Parser}


Our enhanced rule-based aligner produces multiple alignments, and
we would like to use our parser to evaluate their qualities.
A parameterized parser does not accomplish such goal because
training its parameters depends on the aligner's outputs.
A deterministic parser works in this situation but is required to consider
the association between concepts and spans.
This stops the deterministic parsers which build AMR graph only from
the derived concepts\footnote{e.g. the reference relation identifier in \citet{flanigan-EtAl:2014:P14-1}
	and the oracle transducer in \citet{wang-xue-pradhan:2015:NAACL-HLT}.}
from being used because
they do not distinguish alignments that yields to the same set of concepts.\footnote{recall the ``\textit{nuclear}'' example in Section \ref{sec:intro}.}
\begin{table*}[t]
	\small
	\centering
	\begin{tabular}{rllp{16em}}
		Transition & Current State &  Resulting State & Description\\
		\hline
		{\sc Drop}
		& $[\sigma | \texttt{s}_0,\ \delta,\ b_0 | \beta,\ A]$ 
		& $[\sigma | \texttt{s}_0,\ \delta,\ \beta,\ A]$ 
		& pops out the word that doesn't convey any semantics (e.g., function words and punctuations).\\[0.5ex]
		\hdashline
		{\sc Merge}
		& $[\sigma | \texttt{s}_0,\ \delta,\ b_0 | b_1 | \beta,\ A]$ 
		& $[\sigma | \texttt{s}_0,\ \delta,\ b_0\text{\_}b_1 | \beta,\ A]$ 
		& concatenates a sequence of words into a span, which can be derived as a named entity ({\tt name}) or {\tt date-entity}. \\[0.5ex]
		\hdashline
		{\sc Confirm({\tt c})} 
		& $[\sigma | \texttt{s}_0,\ \delta,\ b_0 | \beta,\ A]$
		& $[\sigma | \texttt{s}_0,\ \delta,\ \texttt{c} | \beta,\ A]$ 
		& derives the first element 
		of the buffer (a word or span) into a concept {\tt c}.\\[0.5ex]
		\hdashline
		{\sc Entity({\tt c})} 
		& $[\sigma | \texttt{s}_0,\ \delta,\ b_0 | \beta,\ A]$
		& $[\sigma | \texttt{s}_0,\ \delta,\ \texttt{c} | \beta,\ A \cup \text{relations}(c)]$ 
		& a special form of {\sc Confirm}
		that derives the first element into an entity and
		builds the internal entity AMR fragment.\\[0.5ex]
		\hdashline
		{\sc New({\tt c})} 
		& $[\sigma | \texttt{s}_0,\ \delta,\ \texttt{b}_0 | \beta,\ A]$ 
		& $[\sigma | \texttt{s}_0,\ \delta,\ \texttt{c} | \texttt{b}_0 | \beta,\ A]$ 
		& generates a new concept {\tt c} and pushes it to
		the front of the buffer.\\[0.5ex]
		\hdashline
		{\sc Left({\tt r})} 
		& $[\sigma | \texttt{s}_0,\ \delta,\ \texttt{b}_0 | \beta,\ A]$ 
		& $[\sigma | \texttt{s}_0,\ \delta,\ \texttt{b}_0 | \beta,\ A\cup \{ \texttt{s}_0\xleftarrow{\texttt{r}}\texttt{b}_0 \}]  $
		&  \multirow{2}{15em}{links a relation {\tt r} between the top concepts on
			the stack and the buffer.} \\
		{\sc Right({\tt r})} 
		& $[\sigma | \texttt{s}_0,\ \delta,\ \texttt{b}_0 | \beta,\ A]$ 
		& $[\sigma | \texttt{s}_0,\ \delta,\ \texttt{b}_0 | \beta,\ A\cup \{ \texttt{s}_0\xrightarrow{\texttt{r}}\texttt{b}_0 \}]  $
		& \\[1.5ex]
		\hdashline
		{\sc Cache} 
		& $[\sigma | \texttt{s}_0,\ \delta,\ \text{b}_0 | \beta,\ A]$ 
		& $[\sigma,\ \texttt{s}_0 | \delta,\ \text{b}_0 | \beta,\ A]$ 
		& passes the top concept of the stack onto the deque. \\[0.5ex]
		\hdashline
		{\sc Shift} 
		& $[\sigma | \texttt{s}_0,\ \delta,\ \texttt{b}_0 | \beta,\ A]$
		& $[\sigma | \texttt{s}_0 | \delta | \texttt{b}_0,\ [\ ],\ \beta,\ A]$ 
		&  shifts the first concept of the buffer onto the stack along
		with those on the deque.\\[0.5ex]
		\hdashline
		{\sc Reduce}
		& $[\sigma | \texttt{s}_0,\ \delta ,\ \text{b}_0 | \beta,\ A]$ 
		& $[\sigma,\ \delta,\ \text{b}_0 | \beta,\ A]$
		& pops the top concept of the stack. \\
	\end{tabular}
	\caption{The transition system. 
		The letters in {\tt monospace} font represent the concepts,
		the {\it italic} letters represent the word, and the 
		letters in normal font are either concepts or words.}\label{tbl:trans-sys}
\end{table*}

This discussion shows that to evaluate the quality of
an alignment, we need a deterministic (oracle) parser which
builds the AMR graph from the raw sentence.
\citet{ballesteros-alonaizan:2017:EMNLP2017} presented a 
transition-based parser that directly parses a sentence into its AMR graph.
A transition system
which extends the swap-based dependency parsing system
to handle AMR non-projectivities \cite{damonte-cohen-satta:2017:EACLlong}
was proposed in their work.
Their work presented the possibility for the oracle parser,
but their oracle parser was not touched explicitly.
What's more, in the non-projective dependency parsing,
\citet{choi-mccallum:2013:ACL2013}'s extension to the list-based system \cite{nivre2008algorithms}
with caching mechanism achieves expected linear time complexity and
requires fewer actions to parse a non-projective tree than the swap-based system.
Their extension to transition-based AMR parsing is worth studying.

In this paper, we propose to extend \citet{choi-mccallum:2013:ACL2013}'s transition system to AMR parsing
and present the corresponding oracle parser.
The oracle parser is used for tuning our aligner and training our parser.
We also present a comprehensive comparison of our system with that of 
\citet{ballesteros-alonaizan:2017:EMNLP2017} in Section \ref{sec:comp-ba17}.

\subsection{List-based Extension for AMR Parsing}\label{sec:trans-sys}



We follow \citet{choi-mccallum:2013:ACL2013} and define a state in our transition
system as a quadruple ${ s = (\sigma, \delta, \beta, A) } $, 
where $\sigma$ is a stack holding processed words, $\delta$ is a deque holding words
popped out of $\sigma$ that will be pushed back in the future, and $\beta$
is a buffer holding unprocessed words. $A$ is a set of labeled relations.
A set of actions is defined to parse sentence into AMR graph.
Table \ref{tbl:trans-sys} gives a formal illustration of these actions and how they work.
The first five actions in Table \ref{tbl:trans-sys} are our extended actions, and
they are used to deriving concepts from
the input sentence.

\subsection{Oracle Parser}\label{sec:oracle-parser}

Given an alignment and the gold standard AMR graph, we can build
the best AMR graph by repeatedly applying one of these actions and this is what we called {\it oracle} parser.
Before running the oracle parser, we first remove the
concepts which aren't aligned with any span of words from the AMR graph.
During running the oracle parser,
for a state ${ s = (\sigma | \texttt{s}_0,\ \delta,\ \text{b}_0 | \text{b}_1 | \beta,\ A) } $, 
our oracle parser decides which action to apply by checking the following conditions one by one.
\begin{enumerate}
	\setlength\itemsep{0em}
	\item If $\text{b}_0$ is a word and it doesn't align to any concept, perform {\sc Drop}.
	\item If $\text{b}_1$ is within a span in the alignment, perform {\sc Merge}.
	\item If $\text{b}_0$ is a word or span and it only aligns to one entity concept {\tt c}, perform {\sc Entity({\tt c})}.
	\item If $\text{b}_0$ is a word or span and it aligns to one or more concepts, perform {\sc Confirm({\tt c})} where
	{\tt c} is the concept $\text{b}_0$ aligns and has the longest graph distance to the root.
	\item If $\text{b}_0$ is a concept and its head concept $c$ has the same alignment as $\text{b}_0$,
	perform {\sc New({\tt c})}.
	\item If $\text{b}_0$ is a concept and there is an unprocessed edge {\tt r} between $\texttt{s}_0$ and $\texttt{t}_0$,
	perform {\sc Left({\tt r})} or {\sc Right({\tt r})} according to {\tt r}'s direction.
	\item If $\texttt{s}_0$ has unprocessed edge, perform {\sc Cache}.
	\item If $\texttt{s}_0$ doesn't have unprocessed edge, perform {\sc Reduce}.
	\item perform {\sc Shift}.
\end{enumerate}

We test our oracle parser on the {\it hand-align} data created by \citet{flanigan-EtAl:2014:P14-1}
and it achieves 97.4 Smatch F1 score.\footnote{
	Since some alignments in {\it hand-align} were created on incorrect AMR annotations, we filter out
	them and only use the correct subset which has 136 pairs of alignment and AMR graph.
	This data is also used in our intrinsic evaluation.}
Besides the errors resulted from incorrect manual alignments, entity errors made by the limitation of
our  {\sc Entity({\tt c})} action count a lot.
Since our {\sc Entity} action directly converts
the surface form of a word span into an entity. 
It cannot correctly generate 
entity names when they require derivation,\footnote{e.g., 
``{\it North Koreans}'' cannot be parsed into {\tt  (name :op1 "North" :op2 "Korea")}}
or where tokenization errors exist.\footnote{e.g., 
``{\it Wi Sung - lac}'' cannot be parsed into {\tt (name :op1 "Wi" :op2 "Sung-lac")} }

\subsection{Tune the Aligner with Oracle Parser}\label{sec:tune}

\begin{figure}[t]
	\centering
	\small
	\begin{tikzpicture}
	\node[rectangle,rounded corners,draw=black] (aligner) {Aligner};
	\node[rectangle, draw=black, text width=1.1cm] (input) [left=0.5cm of aligner] {Training Data};
	\node (dot) [right=0.2cm of aligner] {$\vdots$};
	\node[above of=dot] (a1){a$_1$};
	\node[below of=dot] (an){a$_n$};
	\node[rectangle,minimum height=2.7cm,rounded corners,draw=black,right=0.3cm of dot] (oracle){Oracle};
	\node[right=1.5cm of a1] (g1){g$_1$};
	\node[below of=g1] (dot2){$\vdots$};
	\node[right=1.5cm of an] (gn){g$_n$};
	\node[rectangle,minimum height=2.7cm,rounded corners,draw=black,right=0.3cm of dot2] (eval){Eval.};
	\node[right=1.2cm of g1] (s1){s$_1$};
	\node[below of=s1] (dot3){$\vdots$};
	\node[right=1.2cm of gn] (sn){s$_n$};
	\draw [->] (aligner.north) -- (a1.west);
	\draw [->] (aligner.south) -- (an.west);
	\draw [->] (a1.east) -- (g1.west);
	\draw [->] (an.east) -- (gn.west);
	\draw [->] (g1.east) -- (s1.west);
	\draw [->] (gn.east) -- (sn.west);
	\draw [->] (sn.south) to [bend left=30] (an.south);
	\draw [->] (input.east) -- (aligner.west);
	\node [below of=gn] (pick) {highest-scored, pick};
	\end{tikzpicture}
	\caption{The workflow of tuning the aligner with the oracle parser.
	$a_i$ denotes the $i$-th alignment,
	$g_i$ denotes the $i$-th AMR graph, and 
	$s_i$ denotes the score of the $i$-th AMR graph.}\label{fig:workflow}
\end{figure}
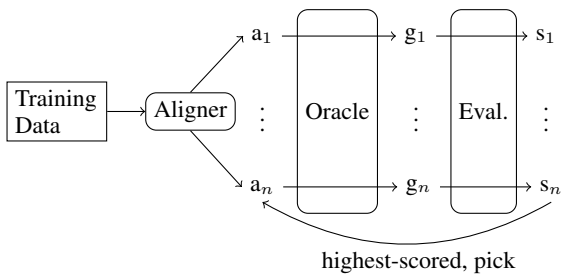
Using our oracle parser, we tune the aligner by picking the alignment 
which leads to the highest-scored AMR graph from the set of candidates
(see Figure \ref{fig:workflow} for the workflow).
When more than one alignment achieve the highest score, we choose the one
with the smallest number of actions.
Intuitively, choosing the one with the smallest number of actions will
encourage structurally coherent alignment\footnote{e.g. the first ``\textit{nuclear}'' aligned to \texttt{nucleus\url{~}1} in Fig. \ref{fig:eg_amr}}
because coherent alignment requires fewer \textsc{Cache} actions.

\subsection{Parsing Model}\label{sec:parser}
Based on our aligner and transition system, we propose a transition-based parser
which parse the raw sentence directly into its AMR graph.
In this paper, we follow \citet{ballesteros-alonaizan:2017:EMNLP2017}
and use StackLSTM \cite{dyer-EtAl:2015:ACL-IJCNLP} to model the states.
The score of a transition action $a$ on state $s$ is calculated as
\[
p(a|s)=\frac{\exp\{g_a\cdot \textsc{StackLSTM}(s) + b_a\}}{\sum_{a'}\exp\{g_{a'}\cdot \textsc{StackLSTM}(s) + b_{a'}\}}\text{,}
\]
where \textsc{StackLSTM}$(s)$ encodes the state $s$ into a vector
and $g_a$ is the embedding vector of action $a$.
We encourage the reader to refer \citet{ballesteros-alonaizan:2017:EMNLP2017}
for more details.

\paragraph{Ensemble.}
Ensemble has been shown as an effective way of improving the neural model's performance \cite{he-EtAl:2017:Long3}.
Since the transition-based parser directly parse a sentence into its AMR graph,
ensemble of several parsers is easier compared to the two-staged AMR parsers.
In this paper, we ensemble the parsers trained with different initialization
by averaging their probability distribution over the actions.

\section{Alignment Experiments}\label{sec:align-exp}

\subsection{Settings}

We evaluate our aligner on the LDC2014T12 dataset.
Two kinds of evaluations are carried out including the {\it intrinsic} and {\it extrinsic} evaluations.

For the {\it intrinsic} evaluation, we follow \citet{flanigan-EtAl:2014:P14-1}
and evaluate the F1 score of the alignments produced by our aligner
against the manually aligned data created in their work
({\it hand-align}).
We also use our oracle parser's performance as an {\it intrinsic} evaluation
assuming that better alignment leads to higher scored oracle parser.

For the {\it extrinsic} evaluation, we plug our alignment into two open-sourced AMR parsers: 1) JAMR \citep{flanigan-EtAl:2014:P14-1,flanigan-EtAl:2016:SemEval}
and 2) CAMR \cite{wang-xue-pradhan:2015:NAACL-HLT,wang-xue-pradhan:2015:ACL-IJCNLP} and evaluate the final performances of the AMR parsers
on both the newswire proportion and the entire dataset of LDC2014T12.
We use the configuration in \citet{flanigan-EtAl:2016:SemEval} for JAMR 
and the configuration in \citet{wang-xue-pradhan:2015:ACL-IJCNLP} without 
semantic role labeling (SRL) features for CAMR.

\subsection{Results}

\begin{table}[t]
	\centering
	\begin{tabular}{lcc}
		\hline
		Aligner & Alignment F1 & Oracle's Smatch\\
		& (on hand-align) & (on dev. dataset)\\
		\hline
		JAMR & 90.6 & 91.7 \\
		Our & 95.2 & 94.7 \\
		\hline
	\end{tabular}
	\caption{The intrinsic evaluation results.}\label{tbl:int-eval}
\end{table}

\paragraph{Intrinsic Evaluation.}
Table \ref{tbl:int-eval} shows the intrinsic evaluation
results, in which our alignment
intrinsically outperforms JAMR aligner by
achieving better alignment F1 score and leading
to a higher scored oracle parser.

\paragraph{Extrinsic Evaluation.}

\begin{table}[t]
	\centering
	\begin{tabular}{lcc}
		\hline
		model & newswire & all \\
		\hline
		\multicolumn{2}{l}{JAMR parser: Word, POS, NER, DEP} & \\
		\quad + JAMR aligner & 71.3 & 65.9 \\
		\quad + Our aligner & 73.1 & 67.6 \\
		\hdashline
		\multicolumn{2}{l}{CAMR parser: Word, POS, NER, DEP} & \\
		\quad + JAMR aligner & 68.4 & 64.6 \\
		\quad + Our aligner & 68.8 & 65.1 \\
		\hline
	\end{tabular}
	\caption{The parsing results.
	}\label{tbl:ext-eval}
\end{table}

Table \ref{tbl:ext-eval} shows the results.
From this table, we can see that our alignment
consistently improves all the parsers by a margin ranging
from 0.5 to 1.7.
Both the intrinsic and the extrinsic evaluations show the effectiveness
our aligner.

\subsection{Ablation}

\begin{table}[t]
	\centering
	\begin{tabular}{lc}
		\hline
		model & newswire  \\
		\hline
		JAMR parser + Our aligner & 73.1 \\
		\quad - Semantic matching & 72.7 \\
		\quad - Oracle Parser Tuning & 67.6 \\
		JAMR parser + JAMR aligner & 71.3 \\
		\hline
	\end{tabular}
	\caption{The ablation test results.}\label{tbl:abl-test}
\end{table}

To have a better understanding of our aligner, we conduct
ablation test by removing the
{\it semantic matching} and {\it oracle parser tuning} respectively
and retrain the JAMR parser on the newswire proportion.
The results are shown in Table \ref{tbl:abl-test}.
From this table, we can see that removing either of these components
harms the performance. 
Removing {\it oracle parser tuning} leads to severe performance drop
and the score is even lower than that with JAMR aligner.
We address this observation to that alignment noise is introduced by
the semantic matching especially by the word embedding similarity component.
Without filtering the noise by our oracle parser,
just introducing more matching rules will harm the performance.

\section{Parsing Experiments}\label{sec:parse-exp}

\subsection{Settings}
We use the same settings in our aligner extrinsic evaluation
for the experiments on our transition-based parser.
For the input to the parser, we tried two settings: 1) using only words as input,
and 2) using words and POS tags as input.
Automatic POS tags are assigned with Stanford POS tagger \cite{manning-EtAl:2014:P14-5}.
Word embedding from \citet{ling-EtAl:2015:NAACL-HLT} is used
in the same way with \citet{ballesteros-alonaizan:2017:EMNLP2017}.
To opt out the effect of different initialization in training the neural network,
we run 10 differently seeded runs and report their average performance following \citet{reimers-gurevych:2017:EMNLP2017}.

\subsection{Results}
\begin{table}[t]
	\centering
	\begin{tabular}{lcc}
		\hline
		model & newswire & all \\
		\hline
		\multicolumn{2}{l}{Our single parser: Word only} & \\
		\quad + JAMR aligner &  68.6 & 63.9 \\
		\quad + Our aligner &  69.3 & 64.7 \\
		\hdashline
		\multicolumn{2}{l}{Our single parser: Word, POS} & \\
		\quad + JAMR aligner & 68.8 & 64.6 \\
		\quad + Our aligner & 69.8 & 65.2 \\
		\hline\hline
		\multicolumn{2}{l}{Our ensemble: Word only + Our aligner} & \\
		\quad x3 & 71.9 & 67.4 \\
		\quad x10 & 72.5 & 68.1 \\
		\hdashline
		\multicolumn{2}{l}{Our ensemble: Word, POS + Our aligner} & \\
		\quad x3 & 72.5 & 67.7 \\
		\quad x10 & 73.3 & \bf 68.4 \\
		\hline\hline
		BA17: Word only$\dagger$ & 68 & 63 \\
		\quad + POS & 68 & 63 \\
		\quad + POS, DEP & 69 & 64 \\
		\citet{damonte-cohen-satta:2017:EACLlong}$\ddagger$ & - & 66 \\
		\hdashline
		\citet{artzi-lee-zettlemoyer:2015:EMNLP} & 66.3 & - \\
		\citet{wang-xue-pradhan:2015:ACL-IJCNLP} & 70 & 66 \\
		\citet{pust-EtAl:2015:EMNLP} & - & 67.1 \\
		\citet{zhou-EtAl:2016:EMNLP20163} & 71 & 66 \\
		\citet{goodman-vlachos-naradowsky:2016:P16-1} &70 & - \\
		\citet{wang-xue:2017:EMNLP2017} & - & 68.1 \\
		\hline		
	\end{tabular}
	\caption{The parsing results.
		x$n$ denotes the ensemble of $n$ differently initialized parsers.
		The difference in rounding is due to previous works
		report differently rounded results. $\dagger$ BA17
		represents the result of \citet{ballesteros-alonaizan:2017:EMNLP2017}, 
		$\ddagger$
		\citet{damonte-cohen-satta:2017:EACLlong}'s result
		is drawn from \citet{ballesteros-alonaizan:2017:EMNLP2017}.
	}\label{tbl:exp-result}
\end{table}

Table \ref{tbl:exp-result} shows the performance of our transition-based parser
along with comparison to the parsers in the previous works.
When compared with our transition-based counterpart
\citep{ballesteros-alonaizan:2017:EMNLP2017}, our word-only
model outperforms theirs  using the same JAMR alignment.
The same trend is witnessed using words and POS tags as input.
When replacing the JAMR alignments with ours,
the parsing performances are improved in the same way as in Table \ref{tbl:ext-eval},
which further confirms the effectiveness of our aligner.

The second block in Table \ref{tbl:exp-result} shows the results of our ensemble parser,
in which ensemble significantly improves the performance and more parsers ensembled, more improvements are achieved.
An ensemble of 10 parsers with only words as input
achieves 68.1 Smatch F1 score which is comparable to the 
AMR parser of \citet{wang-xue:2017:EMNLP2017}.
Using the minimal amount of additional syntactic information -- POS tags,
the performance of the ensemble of 10 parsers
is further pushed to 68.4, which surpasses that of \citet{wang-xue:2017:EMNLP2017} 
which relied on named entity recognition (NER) and dependency parsing (DEP).

A further study on the speed shows that our 10 parser ensemble
can parse 43 tokens per second which
is faster than JAMR (7 tokens/sec.) and CAMR (24 tokens/sec.)
thanks to the simplicity of our model and independence of preprocessing,
like NER and DEP.\footnote{In our speed comparison, we also count the time of preprocessing for JAMR and CAMR. 
	All the comparison is performed in the same single-threaded settings.}

\subsection{Comparison to \citet{ballesteros-alonaizan:2017:EMNLP2017}}\label{sec:comp-ba17}

To explain the improved performance against \citet{ballesteros-alonaizan:2017:EMNLP2017} in Table \ref{tbl:exp-result},
we give a comprehensive comparison between
our transition system and that of \citet{ballesteros-alonaizan:2017:EMNLP2017}.

\paragraph{Capability.} In both these two systems, a span of words can only be derived into concept for one time.
``Patch'' actions are required to generate new concepts from the one that 
is aligned to the same span.\footnote{
	e.g., three concepts in
	the fragment {\tt (person :source (country :name (name :op1 "North" :op2 "Korea")))}
	are aligned to ``{\it North Koreans}''.}
\citet{ballesteros-alonaizan:2017:EMNLP2017} uses a {\sc Dependent} action
to generate one tail concept for one hop and cannot deal with the cases
which have a chain of more than two concepts aligned to the same span.
Our list-based system differs theirs by using a {\sc New} action to deal these cases.
Since the new concept is pushed onto the buffer, {\sc New} action can be repeatedly
applied and used to generate arbitrary concepts that aligned to the same span. 
On the development set of LDC2014T12, our oracle achieves 91.7 Smatch F1 score
over the JAMR alignment, which outperforms \citet{ballesteros-alonaizan:2017:EMNLP2017}'s oracle 
(89.5 in their paper) on the same alignment.
This result confirms that our list-based system is more
powerful.

\begin{figure}
	\includegraphics[width=\columnwidth]{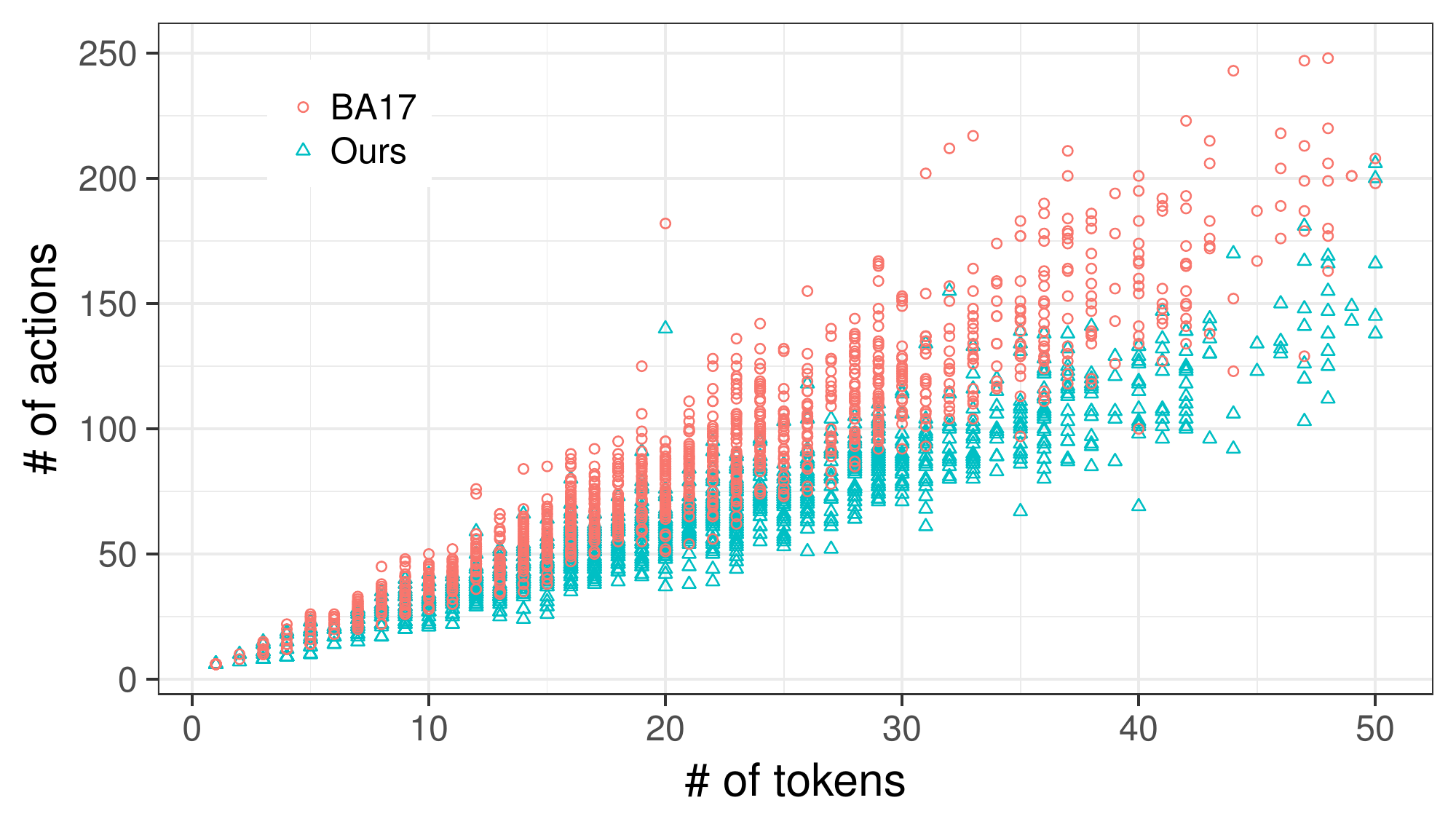}
	\caption{Number of actions required to 
		parse the development set by two systems.}\label{fig:tok_act}
\end{figure}

\paragraph{Number of Actions.} Our list-based system also differs theirs in the number of
oracle actions required to parse the same AMR graphs.
We use the oracles from two systems to parse the development set of LDC2014T12
on the same JAMR alignments.
Figure \ref{fig:tok_act} shows the comparison in which our system
clearly uses fewer actions (the average number of our system is 63.7 and that of
\citet{ballesteros-alonaizan:2017:EMNLP2017} is 86.4).
Using fewer actions makes the parser learned from the oracle less
prone to error propagation. 
We attribute the improved performance in Table \ref{tbl:exp-result}
to this advantage of transition system.

\section{Conclusion}

In this paper, we propose a new AMR aligner which is tuned by
a novel transition-based AMR oracle parser. 
Our aligner is also enhanced by rich semantic resource
and recalls more alignments.
Both the intrinsic and extrinsic evaluations show the effectiveness
of our aligner
by achieving higher alignment F1 score and
consistently improving two open-sourced AMR parsers.
We also develop transition-based AMR parser
based on our aligner and transition system
and it achieves a performance of 68.4 Smatch F1 score
via ensemble
with only words and POS tags as input.

\section*{Acknowledgments}
We thank the anonymous reviewers for their helpful comments and suggestions.
This work was supported by the National Key Basic Research
Program of China via grant 2014CB340503 and the
National Natural Science Foundation of China (NSFC) via
grant 61632011 and 61772153.

\bibliographystyle{acl_natbib_nourl}

\end{document}